\documentclass[10pt,conference]{IEEEtran}
\usepackage[T1]{fontenc}
\usepackage{booktabs,amsmath,amssymb,graphicx,url}
\usepackage[hidelinks]{hyperref}
\title{When Passing Checks Do Not Establish Completion: Evidence Interpretation in Coding Agents}
\author{\IEEEauthorblockN{Xueping Gao}
\IEEEauthorblockA{Alibaba Cloud Computing}}
\begin{document}
\maketitle

\begin{abstract}
Verification procedures in coding agents must preserve execution facts, interpret their scope and discharge checking obligations. We study these responsibilities through qualified repository checks, controlled evidence consumers and executable workflows. A prospective thirteen-task factorial study finds that correct reported facts need not yield the requested finite verdict. Conversely, all 122 delivered labels in a held-out receipt-supply study are correct, while sixteen responses misreport facts and eleven others exceed authorized scope. Production repair supplies successful controls and finite-supported preservation failures. Across 118 repair episodes on 31 tasks, source interventions separate startup reachability from repair and witness production from the claimed operation. Lifetime and resource controls delimit what passing validation establishes. Generated-note comparisons yield local repair differences but can state false explanations; strong ordinary baselines, new tasks and Skill-transfer studies supply no consistent overall advantage. These results characterize evidence interpretation, applicability and fulfilled actions while limiting claims of general diagnostic superiority, Skill deterioration or deployed folding effects.
\end{abstract}

\noindent\emph{Drafting assistance: OpenAI Codex~\cite{aiassist}.}

\section{Introduction}
A coding agent validates changes by executing code from the repository it modifies. That production code may run discovery, command dispatch or output capture, or supply a test's recorded events. Program comprehension therefore requires tracing the edit through validation to completion: did checking reach the intended operation, and did that operation produce the evidence used to claim completion?

Repair research distinguishes plausible from correct patches~\cite{qiplausibility,smithoverfit,ahmedoverfit}. We examine the path from a production edit through validation to completion. That path can change while test, assertion and runner remain unchanged. For example, adding flushing to write can satisfy a flush assertion while the requested explicit flush remains ineffective. Explaining this PASS requires identifying the witness's origin.

Moving specialist verification into a main loop and packaging procedures as Skills motivated this study. Anthropic recommends end-to-end migration evaluation~\cite{sarkis}; controlled comparisons find task-dependent coordination benefits~\cite{agentscaling}. Executable Code Knowledge binds code to contracts/evidence~\cite{eck}; Skill Runtime Intelligence reconstructs execution stages~\cite{sri}; compositional routing organizes retrieval/composition~\cite{routing}. We ask what evidence establishes after production edits. The original repair cohorts do not identify folding or evolution effects. Exploratory handoff and Skill studies are retained in the supplement; neither is a core efficacy claim of this paper.

We examine two interfaces in the same verification path: production code and runners generate observations; agents interpret those observations to report completion. RQ1 asks when factual observations support the reported finite judgment and completion scope. RQ2 asks which production/validation dependencies explain whether a finite result supports an operation-level claim. Source interventions address RQ2; generated-note comparisons test a separate practical consequence.

Our contribution is an empirical account of production-to-validation dependencies, separating defect detection from explanations of the validation outcome. Three findings delimit it. \textbf{Facts, verdicts and authorized completion differ:} controlled interpretation assays separate them. \textbf{Production code participates in validating its edits:} reversions identify startup obstruction; witness, lifetime and admission controls separate executed work from the claimed operation or interval. \textbf{Repair success does not certify its explanation:} generated advice yields local differences while making unsupported claims. We distinguish task breadth from within-task interventions and compare practical consequences with ordinary diagnosis. \emph{Drafting assistance: OpenAI Codex~\cite{aiassist}.}

\section{Study design}
\begin{figure*}[t]
\centering
\includegraphics[width=0.98\textwidth]{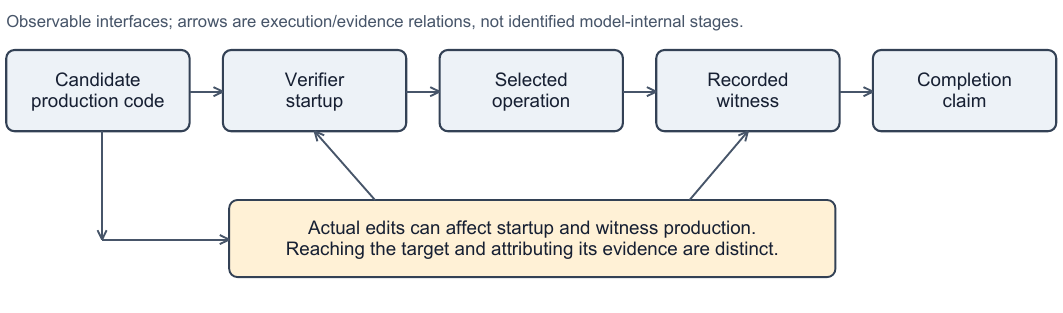}
\caption{Production code participates in both the behavior under test and the production of validation evidence. A candidate can obstruct startup or change which operation supplies a positive witness. The diagram organizes observable interfaces; it is not a model-internal causal graph or an automated integrity guarantee.}\label{fig:production-path}
\end{figure*}
\subsection{Observable units and outcomes}
The unit of analysis is a repository task identified by its benchmark ID or public issue and frozen revision. Distinct IDs are not assumed to be random or statistically independent samples of all defects. Multiple conditions, seeds, source variants and repeated executions within a task are not independent task replications. We distinguish a requested check, actual target entry, its finite outcome, an emitted completion claim and broader behavioral adequacy. A runner error before entry is \texttt{UNKNOWN}, rather than a failing target assertion. For the production-repair cohorts, a finite-supported completion requires at least one accepted edit, a terminal \texttt{finish/fixed} action, a latest PASS receipt for each of the two named checks (changed and companion) bound to the final source digest, and PASS on both post-termination visible checks. Missing receipts, FAIL or UNKNOWN do not satisfy the predicate. Broader official checks and authored contracts are reported separately and do not retroactively replace this finite score.

Records bind revision, source/function identity, phase, registration/body observations, exits and errors. Entry means adapter invocation, not exhaustive coverage. Call stacks distinguish setup and body; file overlap alone does not establish startup execution.

\subsection{Evidence scope and diagnostic logic}
We describe a finite evidence item as $(h,c,p,o,w)$: candidate source digest $h$, selected check $c$, observed phase $p$, relevant operation $o$, and recorded witness $w$. This is an accounting convention rather than a new storage format. The selected check's result remains authoritative only at its executed scope. Concurrent calls require invocation identity; a shared function name does not bind one receipt to every call. Three distinctions drive the analysis:
\begin{enumerate}
\item A requested check and a runner process are not evidence of target entry. We inspect whether the designated invocation and its required assertion-bearing function execute; pre-entry exceptions retain their own status.
\item A positive witness is not automatically evidence of the intended operation. We bind its observed producer and calling site, then use an operation-isolating consumer to test the behavioral consequence.
\item A modified validation context is a different experiment. Replacing startup production code may expose a candidate's target behavior, but the resulting item is not evidence for ordinary execution under the candidate alone.
\end{enumerate}
These distinctions do not turn tracing into whole-program correctness. Original/official-fixed executions qualify the finite defect contrast and regression controls; candidate runs examine the repair. PASS/PASS preservation controls need not distinguish the original defect.

\paragraph{Contract basis and author intervention} A reference contrast alone does not authorize a stronger requirement. Flush forwarding and polling progress come from the reported operations; warning extraction and relation ordering also have unchanged user tests or documented API behavior. Resetting observations, controlling object lifetime and scheduling mtimes are authored diagnostic contexts. They isolate those obligations but do not establish their deployment frequency. We treat annotated choice labels as an explicit application contract, not a universal API requirement. The optional-metadata experiment lacks evidence that its context is supported by Django and therefore supplies only a conditional behavior difference. The focused supplement records these bases and limits separately.

The source-level interventions are narrowly interpretable. A component reversion changes only a recorded fragment of an actual candidate, preserving the ordinary runner and checks. Its causal claim concerns that fragment on the exercised path. Startup-reference substitution changes the path's implementation before restoring the candidate and must carry a conditional interpretation. Passive observation changes neither production source nor the check; outcome reproduction is a gate before interpreting its trace. A failed gate invalidates interpretation. Preservation is necessary, not sufficient: later agent-selected instrumentation preserves a visible result while reading an uninitialized variable. Source-level observation validity requires separate review.

\subsection{Sequential repair cohorts}
Nine cohorts supply 118 repair episodes on 31 tasks (Table~\ref{tab:cohorts}), using Qwen3-Coder-30B native FP8 and Devstral Small~2 24B on L20. Conditions within each cohort share tools, task facts, edit interfaces, per-stage limits and native checks. Study calls exclude synthetic transport fixtures. These are bounded repair workflows, not complete multi-agent reproductions. Exposed exploratory cohorts and later reservations remain separate; their outcomes do not estimate prevalence or a common effect.

\begin{table}[t]
\centering\small
\caption{Actual repair cohorts. Entries are finite-supported completions for Qwen/Devstral; failures and non-completions remain in every denominator. Conditions differ across cohorts; E/F share four task IDs; H/I share three. Counts are not fractions.}\label{tab:cohorts}
\begin{tabular}{lrrl}\toprule
Cohort & Tasks & Episodes & Conditions and Qwen/Devstral\\\midrule
A &4&16& ordinary 1/2; composed 1/2\\
B &6&36& ordinary 0/2; checklist 0/1;\\
&&& composed 0/1\\
C &5&20& ordinary 2/1; dependency 2/2\\
D &3&6& ordinary 2/3\\
E &4&8& ordinary 0/0\\
F &4&8& +check source 1/1\\
G &6&12& ordinary +index 2/3\\
H &3&6& ordinary 1/1\\
I &3&6& larger budget 1/1\\\bottomrule
\end{tabular}
\end{table}

B varies ordinary/checklist/composed instructions; C adds archived-original startup traces. C's input matching is confounded: two ordinary inputs that fit alone are rejected when paired with hypothetical longer inputs. Tokenizer reconstruction reproduces all four affected failures, withdrawing the efficacy inference. All endings remain counted; no Skill-composition/evolution effect is identified. The supplement retains budgets, native-call counts and failed histories.

\subsection{New task qualification and controls}
Source hashes, reservations and exclusions precede actors. Syntax screens nominate mechanisms; original/official-fixed whole-function checks qualify finite defect and preservation contrasts. Five of nine initial reservations qualify, with setup failures retained. Passive traces reproduce 20 outcomes: only one of four changed-definition sets executes before target entry. Reference locations cannot certify candidate footprints.

Later qualifications retain three D tasks, four shared E/F tasks and six G tasks. Source-informed consumers freeze before actors. E has no finite-supported completion; F adds already-visible check source, jointly varying content and length. All five G completions pass prospective consumers; polling diagnosis is post hoc.

A six-candidate screen retains Django16502, Pylint6506 and pytest11160 with three exclusions, following an unsuccessful mock-only screen and source exposure. WSGI iteration/closing~\cite{pep3333}, command help and warning consumers qualify their defects. H/I reuse these tasks/models. Larger limits remove delivery failures but retain warning-only successes; broader unchanged tests reject all twelve endings without incremental consumer alerts. Full qualifications and failures remain in the supplement. \emph{Drafting assistance: OpenAI Codex~\cite{aiassist}.}

\section{RQ1: Facts, scope and completion judgments}
The production analyses extend the registered study of evidence interpretation, applicability and fulfilled checking obligations. Two earlier controlled assays use archived reference receipts to score finite judgments and authorized scope, separately from the repair-completion predicate. Their complete conditions, failures and records are retained in the supplement and anonymous artifact.

A prospective scope-by-rule study qualifies 25 checks on 13 of 16 reserved task blocks: 25 reporters and 400 consumers give 425 calls across four conditions, two models and two dependent seeds. In Qwen3's broad/plain cell, factual fields are correct in 40/50 responses but verdicts in 32/50: eight recover the facts yet abstain incorrectly. Rule effects vary by model; low factual recovery in Qwen2.5 prevents treating its lower fact/verdict disagreement as reliable interpretation.

\begin{figure}[t]
\centering
\includegraphics[width=\columnwidth]{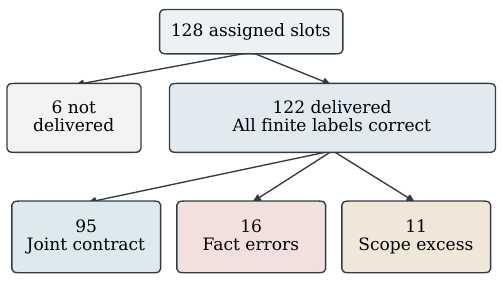}
\caption{Receipt-supply output partition: 16 checks, nine task blocks and two models. All 122 delivered finite labels are correct, but only 95/128 assigned outputs meet the joint fact-and-scope contract. The 16 fact-error outputs may also overclaim; the 11 others have correct facts. Repeated cells are not independent tasks.}\label{fig:receipt-scope}
\end{figure}

A held-out assay crosses complete/missing receipts with issue text present/withheld on 16 checks in nine task blocks and two models. All 122 adapter-delivered labels are correct, but only 95/128 assigned outputs satisfy the joint fact-and-scope contract (Figure~\ref{fig:receipt-scope}). Strict delivery is reported separately. Correct judgment alone therefore leaves an evidence-interpretation obligation. 

 \emph{Drafting assistance: OpenAI Codex~\cite{aiassist}.}

\section{RQ2: Production-to-validation dependencies}
\subsection{Reachability: a Django command-dispatch dependency}
In the six-task cohort, one Devstral composed ending claims a fix after two requested suites return \texttt{UNKNOWN}. Its production edit introduces an unimported name into \texttt{call\_command}; the runner also calls that function before reaching either selected test. Three Qwen endings on the same issue obstruct startup but finish unresolved. Thus there are four candidate endings on one issue, not four independent failures or four unsupported completions.

A phase-local diagnostic runs original, official-fixed and all four candidate sources under normal, sham and startup-reference modes. Normal and sham modes leave all candidates UNKNOWN/UNKNOWN. Substituting the original function only during startup, then restoring the candidate and aliases before setup/body, allows each candidate's two target checks to execute and fail. Source and restored-function hashes bind all 36 outcomes. A serializer-only recovery and the earlier failed fingerprint audit remain in the supplement.

The diagnostic answers a reachability question under changed startup context. It does not show that the unmodified candidate passes or even executes normally. A separately frozen synthetic falsifier makes this limitation concrete: a bootstrap-only fault yields normal UNKNOWN but startup-reference PASS, whereas an all-call fault yields normal UNKNOWN and startup-reference FAIL. These are constructed controls on the same issue, not new natural agent errors. A rescued PASS is therefore conditional diagnostic evidence, not a correctness certificate for ordinary execution.

\subsection{Reachability: an independent pytest obstruction}
In the five-task cohort, a Qwen dependency-condition edit on pytest5262 changes a neighboring \texttt{safe\_text\_dupfile} function and deletes its return value. The requested issue concerns the mode of \texttt{EncodedFile}. The edited neighboring function runs during pytest startup; its missing return produces a \texttt{NoneType} failure before either selected test. All four final checks are UNKNOWN. The actor does not claim completion, and a later request is blocked by the input ceiling. This is an independent repository issue, but not a second unsupported fixed claim or evidence that extra trace information caused the edit.

We revert the two actual edit components independently and execute all four prequalified checks under the ordinary runner (Table~\ref{tab:ablation}). Restoring the return restores execution, exposing the original requested defect. Restoring only the mode leaves startup obstructed. Unlike startup-reference substitution, these are complete candidate variants executed normally. They establish the return deletion's effect on the exercised startup path; they do not turn component variants into independent model attempts.

\begin{table}[t]
\centering\small
\caption{Pytest component reversions: one actual candidate, four whole-function checks.}\label{tab:ablation}
\begin{tabular}{lll}\toprule
Source & Requested check & Three controls\\\midrule
Original & FAIL & PASS\\
Official fixed & PASS & PASS\\
Actual candidate & UNKNOWN & UNKNOWN\\
Return restored & FAIL & PASS\\
Mode restored only & UNKNOWN & UNKNOWN\\\bottomrule
\end{tabular}
\end{table}

Together these issues show why a production edit must be included in the dependency analysis of validation itself. One issue supplies an unsupported ending; the other supplies abstention/non-completion. Neither observation estimates how often this occurs across repositories. \emph{Drafting assistance: OpenAI Codex~\cite{aiassist}.}

\subsection{Witness production: a finite-supported flush repair}
On Django13516, both Qwen conditions produce the same patch: add an unconditional underlying \texttt{flush()} to \texttt{OutputWrapper.write()}, while leaving explicit wrapper \texttt{flush()} ineffective. Both patches pass the canonical changed check and all prequalified additional official checks, and both actors claim a supported fix under the frozen finite predicate. They count as one independent issue and one distinct patch, not replicated independent discoveries. Both Devstral conditions instead implement a guarded explicit flush method and also pass the official checks.

The minimal source distinction is the location of the underlying call. The Qwen candidate appends \texttt{self.\_out.flush()} to \texttt{write}; the successful repairs instead supply a guarded explicit \texttt{flush} that forwards when the underlying output supports it. These are observed production edits, not faults injected for the experiment. In the unchanged fixture, the relevant action order is:
\begin{quote}\ttfamily\small
stdout.write('Working...')\\
stdout.flush()\\
stdout.write('OK')\\
assert underlying\_flush.called
\end{quote}
This excerpt summarizes the recorded fixture order and final predicate; full original whole-function bytes remain in the artifact. The assertion does not reset the mock between writing and explicit flushing. Consequently, an earlier write-side event can satisfy it even when the explicit operation adds none. Our follow-up resets observations at that boundary, without demanding that writes obey a particular buffering policy.

The official fixture writes, explicitly flushes, then writes again; its assertion checks whether the underlying mock flush was called. Passive tracing of the unchanged check preserves all twelve original outcomes across original, official-fixed and four actual sources. The two Qwen witnesses arise from the writes surrounding the explicit flush. No underlying flush event originates at the fixture's explicit-flush line. Official and both Devstral sources have an event from that line (Table~\ref{tab:witness}). Thus the positive boolean alone does not identify which operation fulfilled the obligation.

\begin{table}[t]
\centering\small
\caption{Flush evidence and behavior on one issue. Qwen conditions share one identical patch. Explicit events are attributed to the fixture's flush line; compatibility uses an output object supporting write without flush.}\label{tab:witness}
\begin{tabular}{lcccc}\toprule
Source & Official & Explicit & Explicit & Write-only\\
& check & events & behavior & output\\\midrule
Original & FAIL &0&FAIL&PASS\\
Official fixed &PASS&1&PASS&PASS\\
Qwen, both &PASS&0&FAIL&FAIL\\
Devstral, both &PASS&1&PASS&PASS\\\bottomrule
\end{tabular}
\end{table}

\subsection{Behavioral consequence and counterexamples}
An operation-specific follow-up clears earlier flush observations after a write and then explicitly flushes. Both Qwen patches still fail to forward that operation; official and both Devstral repairs pass. A compatibility control uses a sink with write but no flush, supported by both original and official implementations. The new unconditional write-side call raises an exception for this previously valid output. The result is a concrete new regression, not a disagreement about eager versus lazy output formatting.

The pre-outcome nested-wrapper consumer already distinguishes both Qwen patches from the successful Devstral repairs. Across all three consumer pairs, all twelve relevant actor endings are retained. The Devstral ordinary pytest ending passes its mode-directed consumer but fails the named official mode check and never claims a fix. This is an important counterexample: one consumer property also cannot replace whole-issue evidence. All four runserver endings fail their consumers without claiming completion.

The unchanged selected check gives inadequate Qwen and successful Devstral repairs the same original/candidate/gold signature, (FAIL, PASS, PASS). This supplies defect contrast, not an operation contract. We distinguish three obligations:
\begin{enumerate}
\item \textbf{Live explicit forwarding.} For a live wrapper over a flush-capable sink, reset prior observations after writing, invoke wrapper \texttt{flush()}, and require an underlying flush event within that call, before it returns. Earlier writes and later cleanup cannot satisfy this obligation.
\item \textbf{Cleanup-context forwarding.} When cleanup exercises the wrapper's flush method on a flush-capable sink, that method forwards to the sink in that context. This is method-level evidence, not proof of timely explicit forwarding while the wrapper is live.
\item \textbf{Write-only compatibility.} A previously supported sink with \texttt{write} but no \texttt{flush} must still accept wrapper writes without an added flush-attribute exception.
\end{enumerate}
Let $T(P)$ denote the selected check's outcome triple and $O(P)$ the operation-reset observation for the first obligation. The actual Qwen source $P_q$ and a successful Devstral source $P_d$ satisfy
\[
T(P_q)=T(P_d)=(\mathrm{F},\mathrm{P},\mathrm{P}),\qquad
O(P_q)=0,\quad O(P_d)=1.
\]
No decision using only $T$ can recover $O$ for both sources: equal inputs require equal decisions. This one-pair separation concerns the triple alone, not BSG-VA's richer command capture, replay and evidence-role analysis~\cite{bsg}. It neither refutes that method nor establishes practical breadth.

\subsection{A lifecycle counterexample to callsite gating}
The passive protocol predicted one total flush event for official and Devstral sources, but observed two: one from explicit fixture flush and another without its handler frame. We retain that failed prediction and separately freeze a 42-execution diagnostic: six unchanged production states, six command-fixture contexts and six canonical companion controls. The whole check and assertions stay unchanged; fixture variants preserve its required output marker. All twelve canonical outcomes reproduce the passive study, and every marker assertion remains satisfied.

Table~\ref{tab:lifetime} distinguishes the channels. Bypassing wrapper writes and omitting explicit flush still leaves one flush-method event for official and both Devstral repairs. Holding a strong reference to the wrapper removes that event and changes the same boolean assertion from PASS to FAIL. Retaining explicit flush while holding that reference preserves PASS. In contrast, Qwen's two events depend on wrapper writes. These interventions locate a lifetime-sensitive witness channel in this runtime; they do not identify an exact C-level finalizer entry. Python documents that \texttt{IOBase} destruction calls \texttt{close}, which flushes the stream~\cite{pythonio}; this supports the lifecycle interpretation rather than proving the exact observed call chain.

\begin{table}[t]
\centering\small
\caption{Lifetime diagnostic on the same flush issue. P/F denotes the unchanged assertion outcome. Official and both Devstral states agree; identical Qwen patches agree. These are altered fixture contexts, not new repair episodes.}\label{tab:lifetime}
\begin{tabular}{p{.51\columnwidth}cc}\toprule
Fixture context & Fixed/Dev & Qwen\\\midrule
Canonical & P:2 flush & P:2 write\\
Omit explicit flush & P:1 flush & P:2 write\\
Direct writes, explicit flush & P:2 flush & F:0\\
Direct writes, no explicit flush & P:1 flush & F:0\\
Same, retain wrapper & F:0 & F:0\\
Canonical, retain wrapper & P:1 flush & P:2 write\\\bottomrule
\end{tabular}
\end{table}

Three questions must stay distinct. \emph{Callsite necessity} asks whether removing the fixture's explicit call changes the assertion. \emph{Event attribution} asks which operation and context produced a recorded flush. \emph{Contract satisfaction} asks whether that event meets the specified obligation. Cleanup can support cleanup-context forwarding even when the literal fixture call is unnecessary; it cannot discharge live explicit forwarding. The reset and write-only controls separately establish Qwen's two failures. Literal-callsite gating is a deliberately overstrict rule examined here, not a rule attributed to prior methods.

\subsection{An independent task: snapshot evidence and polling progress}
Devstral's Django11177 repair deduplicates \texttt{snapshot\_files}, passes both named checks and its prospective real-file consumer, and finishes fixed. An additional thread-name check fails on both original and actual sources, identifying an uncovered obligation rather than a new regression. Snapshot cardinality, thread naming and polling progress are distinct properties.

A post-hoc consumer drives the actual \texttt{StatReloader.tick}, snapshot and retained-state code. Real-file mtimes are 100/101/102; controlled clock reads are 100.75/101.75/102.75/103.75. Only enumeration, sleep/clock and the notification boundary are replaced. Original, unchanged Qwen ending and actual Devstral source emit no notifications; the official repair emits 101 and 102. All sources pass timely-mtime and unchanged-file controls. This constructed coarse-mtime schedule is motivated by documented filesystem resolution limits~\cite{pythonstat}; it measures neither deployment frequency nor process restart.

The actual source registers a new path only if its mtime exceeds the previous wall clock. The first observation therefore never enters retained state in the main schedule. Whole-tick replacement restores progress. More narrowly, adding first-observation registration restores both notifications while preserving the comparisons and clock reads. Name-only and comment-sham changes do not (Table~\ref{tab:polling}). Removing the registration recovers actual bytes.

\begin{table}[t]
\centering\small
\caption{One-task polling diagnosis. Notifications are observed on the constructed schedule; expected count 2. Snapshot/thread are unchanged official checks. Full-tick variants reproduce the corresponding registration outcomes; all 60 executions remain retained.}\label{tab:polling}
\begin{tabular}{@{}lccc@{}}\toprule
Source & Snapshot & Thread & Notifications\\\midrule
Original / unchanged Qwen &FAIL&FAIL&0\\
Official fixed &PASS&PASS&2\\
Actual Devstral &PASS&FAIL&0\\
+thread name &PASS&PASS&0\\
+registration &PASS&FAIL&2\\
+registration +name &PASS&PASS&2\\
Registration sham &PASS&FAIL&0\\\bottomrule
\end{tabular}
\end{table}

The 42 full-source/component and 18 minimal registration/name/sham executions preserve all natural sources and controls. They separate an unrelated official failure from a residual polling defect on one successful ending. The earlier snapshot consumer's PASS does not establish this downstream progress obligation; no general correctness claim follows.

\subsection{Evidence consumption: an independent regression}
On pytest11160, both model families replace \texttt{issubclass(w.category, cls)} with \texttt{w.category is cls}, pass their two visible checks and finish fixed. The larger-budget cohort repeats the identical patch. These four histories concern one task and one patch. Broader unchanged tests already reject it.

The downstream user test emits a \texttt{UserWarning} and retrieves it through generic \texttt{recwarn.pop()}. Original and official-fixed sources pass; the actual patch fails retrieval. A diagnosis frozen after H source review but before I outcomes retains both references, all four histories and a comment sham (28 executions). Immediately before extraction, every state contains exactly one warning with the expected category and message. Generic extraction fails on every model history and sham; exact-\texttt{UserWarning} extraction passes throughout. Reverting the single predicate recovers original bytes.

This separates successful evidence capture from a modified consumer's rejection of intact evidence. Failure occurs at extraction, before any queue-state assertion. The contrast explains an ordinary regression alert; it adds neither a defect-detection advantage nor a new subtype-bug category.

\subsection{A primary-key fix changes documented relation ordering}
In cohort F, Qwen repairs descending inherited primary-key ordering and finishes with both requested suites and all six final checks passing. Its edit computes related-model ordering but deletes that expression's return, falling through to the foreign-key column. Django's documented relation-ordering contract still applies~\cite{djangoordering}.

A post-hoc parent-link consumer separates key order from rating order: key1/rating1 versus key2/rating9. Original and official sources select key2; the actual repair selects key1, with SQL changing from rating DESC to relation-key ASC. Swapping ratings conceals the regression. Direct-ID and explicit-rating controls pass.

\begin{table}[t]
\centering\small
\caption{Ordering-task source tradeoff. PK is the original discriminator; relation/direct are unchanged official tests. Natural PK outcomes are archived; component PK and all relation/direct outcomes are newly executed.}\label{tab:ordering}
\begin{tabular}{lccc}\toprule
Source & PK & Relation & Direct ID\\\midrule
Original & FAIL & PASS & PASS\\
Official & PASS & PASS & PASS\\
Actual Qwen & PASS & FAIL & PASS\\
Restore return & FAIL & PASS & PASS\\
Remove identities & PASS & FAIL & PASS\\
Return + selective guard & PASS & PASS & PASS\\\bottomrule
\end{tabular}
\end{table}

Thirty executions retain all four E/F histories, both references and three component variants. Restoring the deleted return recovers relation ordering but reopens the primary-key defect; removing the new identity branches does not recover ordering. Restoring the return plus the official selective primary-key guard satisfies both covered obligations (Table~\ref{tab:ordering}). AST bindings retain the actual edit's components. Eighteen further executions show that unchanged upstream relation-ordering tests also detect the regression while direct-ID controls pass. The source intervention identifies the return's role in this tradeoff, not the model's reason for deleting it. Complete sources, consumers and counts remain in the focused supplement.

\subsection{Values and backend preconditions}
On Django13315, DISTINCT repairs pass the selected checks yet change annotated-label choices from four entries to two, whereas original returns four: this is residual behavior, not a new regression. Real PostgreSQL point-field queries expose a separate equality-operator precondition already documented in the repair history~\cite{django13315}. The supplement retains all source states, SQL and scalar preservation controls.

\subsection{Native C: allocation evidence depends on prior state}
On known zstd2094~\cite{zstd2094}, a real C consumer checks correct decoding and zero configured-allocator requests during the call. Construction, initialization and warmup precede measurement. Five sources cross five contexts (25 calls, one issue, no agent repairs). Original, sham and reference-minus-condition decode an empty frame correctly but request four bytes on first use; after warmup, all pass with zero requests. Full reference and a condition-only transplant request zero in both contexts (Table~\ref{tab:zstd}).

\begin{table}[t]
\centering\small\setlength{\tabcolsep}{3pt}
\caption{Configured decoder allocator bytes requested during each measured call. All shown decodes are correct. Constructor allocation is excluded; these are not whole-lifetime or physical-heap measurements.}\label{tab:zstd}
\begin{tabular}{@{}p{.26\columnwidth}rrr@{}}\toprule
Source & Cold empty & Warm empty & Cold nonempty\\\midrule
Original / sham / rollback &4&0&0\\
Full / minimal fix &0&0&0\\\bottomrule
\end{tabular}
\end{table}

Allocation denial during measurement rejects the three old-condition cold decodes but preserves their warm decodes. Nonempty first-use controls also pass. Static-library/link-map bindings and unchanged source components support the contrast: warm evidence does not establish first-use behavior.

\subsection{Observation and execution boundaries}
Additional qualified cases delimit the account. In Catch2~\cite{catch2723}, enabling successful-assertion reporting changes the same binary's unexpected-throw exit from~1 to~0; sham agrees and the reference exits~1 in every mode (54 processes). Guava~\cite{guava2971} exposes retained state despite a null presence query. These are known, source-informed dependencies, not independent discoveries.

An executable Superpowers helper~\cite{skillhelper2011} fixes one discovery glob while regressing the previously working prefixed form. Seventy-two helper executions and twelve direct npm controls separate discovery, attempted execution and body entry: missing npm can yield an ``all tests clean'' ending without entering either body. This is one artifact under authored failure-injection contexts, not learned-Skill degradation or failure with a properly installed runner. Full states and controls remain in the supplements. \emph{Drafting assistance: OpenAI Codex~\cite{aiassist}.}

\subsection{Concurrent calls: settlement and input delivery}
A source-informed ADK replay~\cite{adk7058,adk7059} exercises real tool dispatch, asynchronous generators, queues and shutdown. Two same-name calls start; the base settles only the latest task and aliases their queues. Four native checks cover same-name shutdown, name-based stopping, a single call and distinct names. The pinned, unmerged proposed reference passes all four with cooperative tool generators; the SDK's timeout does not guarantee settlement of uncooperative tools.

Queue identity alone leaves a consumer obligation. An additional consumer exercises actual \texttt{send\_to\_model} broadcast and tool receives; a provider fixture only acknowledges native activity and tool-result messages. Base delivers one activity to only one same-name call; the proposed reference delivers it once to each. Single-call and distinct-name controls pass. A six-ending source-edit comparison (Section~\ref{sec:generated-advice}) supplies the candidates. This consumer was frozen after that cohort completed but before inspecting its efficacy, and qualified before application to every final source. It does not replace the frozen primary score. None of the six actual candidate endings qualifies under either contract, so the extra check supplies no incremental alert on that candidate set; the source reversions below address a separate explanatory question.

Exact, reversible source edits isolate the dependencies (Table~\ref{tab:live-boundaries}). Reverting only input broadcast preserves all four primary checks, yet loses one receiver. Reverting only shutdown bookkeeping preserves delivery to both but leaves a task pending. A comment sham preserves both. These 21 executions concern three authored variants on one reported issue, not new Agent attempts. Whole-file controls, including an incompatible registry reversion, remain in the supplement. The contrast separates invocation identity, settlement and actual input consumption; it establishes neither remote-model behavior nor migration of subagent reasoning into a main loop.

\begin{table}[t]
\centering\small\setlength{\tabcolsep}{3pt}
\caption{Two same-name ADK calls. Four checks concern shutdown, stopping, single and distinct-name calls, respectively. Delivery measures one activity received by each invocation. Source variants remain within one issue.}\label{tab:live-boundaries}
\begin{tabular}{lccc}\toprule
Source & Four checks & Settled & Delivery\\\midrule
Base &F/F/P/P&1/2&1/2\\
Proposed reference &P/P/P/P&2/2&2/2\\
Old broadcast only &P/P/P/P&2/2&1/2\\
Old shutdown only &F/P/P/P&1/2&2/2\\
Comment sham &P/P/P/P&2/2&2/2\\\bottomrule
\end{tabular}
\end{table}

\subsection{Task admission and shutdown in Rust}
Neon's public \texttt{BackgroundPurges} change~\cite{neon8650,tokio6759} supplies a Rust component replay. The intermediate version reads \texttt{is\_closed()}, then later registers a filesystem purge. A hook pauses the spawner after its open-state read. Shutdown closes the tracker and returns with closed/empty both true, yet purging begins afterward. These facts do not establish the component's stated completion guarantee.

The actual merged version reserves a task token before checking closure and retains it through the work. A one-line early-token control repairs the intermediate path while retaining its dispatch call. Both wait for cleanup; a comment sham preserves premature return. All four variants preserve normal purge, closed-request rejection, existing-work drain and dropped-waiter controls; two uninstrumented controls also pass. These 22 processes cover one reported race under a constructed legal interleaving, using the extracted component, native Tokio and real filesystem deletion. Full-server behavior and prevalence are unmeasured.

\section{What the source-level explanations establish}\label{sec:diagnostic-value}
Ordinary errors already flag startup obstruction; broader unchanged tests detect the ordering and warning regressions. The additional question is explanatory: which dependency determines the observed behavior, and which competing account can be ruled out? Table~\ref{tab:diagnostic-value} summarizes contrasts from the completed cases, not predictions made on unseen repairs.

\begin{table*}[t]
\centering\small\setlength{\tabcolsep}{4pt}
\caption{What ordinary checking detects and what the interventions additionally explain. Flush and flush-lifecycle rows concern the same issue. Every conclusion concerns the recorded source, checks and context; source variants are not independent repairs.}\label{tab:diagnostic-value}
\begin{tabular}{@{}p{.12\textwidth}p{.21\textwidth}p{.34\textwidth}p{.27\textwidth}@{}}\toprule
Case & Ordinary evidence & Separating intervention & Supported conclusion and boundary\\\midrule
Pytest startup & Runner errors before target entry & Restore deleted return: startup recovers, target FAIL, controls PASS & Return deletion obstructs this startup path; recovery alone is not repair\\
Flush & Named boolean assertion PASS & Reset before explicit flush: FAIL; trace locates write-origin events & Writes supply the positive witness; no explicit live forwarding in that context\\
Flush lifecycle & Named boolean assertion PASS & Bypass writes, retain wrapper: FAIL; explicit flush retained: PASS & Cleanup is a separate witness channel; not proof of the exact finalizer chain\\
Polling & Thread-name check and notification check disagree & Name-only: 0 notifications; registration-only: 2 while name check still fails & Registration controls progress on the constructed schedule; no deployment-frequency claim\\
Warning & Broader unchanged extraction test FAIL & Inspect intact warning, then substitute exact-category lookup: PASS & Capture succeeds; the modified generic consumer rejects the category\\
Relation ordering & Broader unchanged ordering test FAIL & Remove new branches: FAIL; restore return: ordering PASS, PK defect reopens & The return expression mediates this tradeoff; no incremental defect-detection claim\\
Concurrent tools & Four primary native checks PASS & Revert broadcast only: one input consumer misses; revert shutdown only: one task remains & Settlement and input consumption are distinct obligations on one reported issue\\
Rust purge & Authored closed/empty check PASS & Pause after open-state read; add only an early token: shutdown waits & Snapshot covers registered work; reservation bridges the admission interval\\\bottomrule
\end{tabular}
\end{table*}

\paragraph{Detection, explanation and repair differ} Polling's official failure and notification failure change independently; another failing test alone would not establish why. Inspecting the warning before extraction rules out failed capture despite an ordinary test alert. These are source-level explanations, not measured diagnosis-time or repair-success gains.

\paragraph{Counterexamples constrain the account} Cleanup execution legitimately supports cleanup, but not timely live forwarding. A consumer PASS need not resolve the issue: a non-completing pytest repair passes the mode consumer but fails its named check. Rating alignment hides the ordering regression. No single signal yields an exhaustive taxonomy or universal gate.

A C-library factorial extends the operation distinction: json-c lookup finds a decoded member, but assignment and deletion consume its encoded key differently. Setter-only decoding passes 144/144 assignment contexts but 4/12 removal contexts; removal-only decoding gives 36/144 and 12/12; combined decoding passes both. Original/comment-sham give 36/144 and 4/12. All five sources pass 28 upstream tests. These investigator interventions isolate two operations within one library, not autonomous repair gains or independent task replications; the supplement records source bindings and controls.

An actual generated pugixml comparator separates three obligations: a successful sorting example, a valid comparison relation, and that relation matching current tree order. Its visible example passes, but an incremented counter makes both opposite comparisons true and non-repeatable; protected sorting crashes. Removing only the counter restores the exercised relation laws and removes the crash, yet only 8/28 moved-tree contexts pass. The official reference passes 28/28. Seven source states and 28 native processes bind this post-hoc intervention; it explains one actual repair, without establishing autonomous diagnostic benefit.

We summarize the analyses with four questions: (1) what operation, preconditions, context and interval define the obligation; (2) what does the source-bound existing check actually establish; (3) which producer, consumer or retained-state intervention separates the competing explanations while preserving controls; and (4) is the result a residual defect, a new regression or only an altered-context difference? These questions organize retrospective case analysis. Prospective diagnostic superiority and gains over ordinary full-suite testing remain unestablished. \emph{Drafting assistance: OpenAI Codex~\cite{aiassist}.}

\section{Public scope calibration and successful controls}
Two preselected SWE-agent submissions~\cite{sweagentartifacts} supply twelve artifact/task slots: seven production-only patches qualify and five remain excluded. All 91 native executions and successful controls are retained. Django14315 passes PostgreSQL-subclass checks while the inherited client contract fails. Django15252 passes mocked caller routing while real migration checks fail. Their four public reports already say unresolved: these are scope explanations, not new evaluator false positives. Django16116 and Pylint6528 pass all selected checks; pytest5840 preserves five controls but fails its defect discriminator. Django13112 has an ordinary syntax obstruction. The supplement retains every slot and exclusion.

A separate stream of 2,283 predictions for 2,266 tasks yields 35 syntactic stand-in nominations; five source-informed cases qualify. Frozen scripts pass on original, reference, candidate and comment-sham sources, but real obligations pass only with the reference; preservation controls pass throughout (45 PASS/15 FAIL/0 UNKNOWN). Sphinx10481 and Django14559 voluntarily submit after simulation success. Three other actors end at their budget limit. Crucially, Django15666 imports actual Django and enters ORM representation, but mocked fetching supplies the cache without executing the required SQL-ordering compiler. This rules out import or incidental entry as sufficient evidence. Direct real checks also reject all five candidates, so no superiority to that baseline follows. These added-script cases are separate from the production-only cohorts. \emph{Drafting assistance: OpenAI Codex~\cite{aiassist}.}

\section{Secondary test: generated advice and repair}\label{sec:generated-advice}
Our main contribution is the source-level analysis of validation dependencies. We separately test whether generated interpretation helps repair: explanation accuracy, repair success and evidence scope are different outcomes. All comparisons below are exploratory; repeated seeds and contexts do not increase the independent task count. The benefit supplement retains the complete earlier comparisons, failures and costs, including handoff and Skill studies.

\paragraph{Ordinary diagnosis is a necessary baseline}
Across three new C/C++/Rust tasks, fresh ordinary and operation-specific notes each repair 0/3 versus withholding 1/3. A later equal-input-length re-audit yields operation 1/6, ordinary 2/6 and withholding 0/6 across two seeds. Supplying all six actual helpers per task, notes and visible feedback equally yields 2/3 for ordinary integration analysis, operation analysis and no extra analysis. These contrary results prevent interpreting added analysis as a general repair improvement.

\paragraph{Useful repair can coexist with false explanation}
An agent-selected trace workflow first produces no usable final plan. Mechanical recovery and source auditing remove invalid watches, including pre-initializer reads that preserved visible output. Two fixed seeds then yield trace-plus-generated-note 2/6 versus 0/6 for source-only, raw-trace and trace-withheld-note controls under equal initial repair lengths. Both gains are Rust, passing 112 checks in each build; C/C++ remain unrepaired. A new nlohmann/json task~\cite{nlohmann5693} yields ordinary and operation notes 2/2 successful seeds versus withholding 1/2, each success passing 32 contexts. The notes incorrectly blame the returned iterator: a separate replay finds original iterator equality 32/32 while array equality is 20/32. Thus repair success does not validate every statement in a diagnosis.

\begin{table}[t]
\centering\small
\caption{Content ablation on two adaptively selected tasks, each with two repair seeds. Four existing Agent-generated notes are reused. Replay preserves generated code and changes only extraction; it is not a rerun of the interaction.}\label{tab:benefit-summary}
\begin{tabular}{lcc}\toprule
Transferred content & Original workflow & Source replay\\\midrule
Compatible-repair section & 4/4 & 4/4\\
Full generated note & 1/4 & 2/4\\
No note & 0/4 & 1/4\\\bottomrule
\end{tabular}
\end{table}

\paragraph{Selected-task content ablation}
A two-task component ablation transfers either each actual note, its exact compatible-repair section, or no note, with matched initial lengths and two fresh repair seeds (Table~\ref{tab:benefit-summary}). Uniform exact-code recovery scans all 24 responses and replays all four marker-rejected functions. The section-only advantage survives this sensitivity analysis. However, both tasks were selected after earlier local gains; notes were reused rather than freshly generated. Removing four sections jointly changes salience, potentially misleading information and presentation. This experiment identifies a local content-transfer effect, not why it occurs or whether explanatory reasoning is necessary.

\paragraph{Replication and compatibility limit the local benefit}
Fresh more-itertools notes~\cite{peek1322} yield no repair in six endings on 39 contexts; an exposed repeat also fails all six. Scoped notes confuse reverse with positive slicing. Cachetools supplies no full repair. Boltons~\cite{boltons610} initially favors scoped notes, but a fixed three-seed replication reverses the complete-note direction; uniform source recovery equalizes it at 2/3. Bidict~\cite{bidict377} rejects all eight primary endings; three reach 38/39 while breaking ordinary equality. These trials limit the selected-task contrast. Full per-condition outcomes, failed histories, source-recovery rules and costs remain in the benefit supplement.

\paragraph{Prospective controlled diagnostic decisions}
Four previously unused tasks in attrs (Python), cJSON (C), Gson (Java) and SmallVec (Rust) supply a frozen comparison with two seeds per task. Each arm generates its own explanation, selects one authored candidate for three native checks, then selects a candidate and predicts all nine candidate/check outcomes. Initial input lengths, model-call and output ceilings match. Ordinary/scoped procedures both select the qualified candidate in 6/8 endings, but satisfy the joint choice-and-prediction contract in 3/8 versus 0/8. All sixteen final JSON outputs are valid; each arm has one invalid first-stage probe selection. Four SmallVec outputs contradict their own executed receipt, two per arm. These constrained decisions provide fresh counterevidence to diagnostic superiority, not autonomous repairs. A development follow-up retains all sixteen endings: automatic receipt-conflict feedback and strong ordinary second review both satisfy the joint contract in 6/16 revisions. Feedback reduces receipt contradictions but leaves one malformed output and no overall diagnostic gain. Full inputs, controls, costs and limitations are in the supplementary update.

\paragraph{Native source-edit workflows}
Two navigation workflows share a fresh Agent-generated probe and receipt across newly generated ordinary-note, operation-note and no-note conditions. Six four-file ADK endings yield 0/2 full source passes per arm under matched 12-decision/12-process ceilings. Eighteen endings across ADK cancellation, Tenacity iteration and AnyIO process close yield 2/6,1/6 and 0/6 under matched six-decision/12-process ceilings; all successes concern ADK cancellation. None delivers a finish action, and 14/24 leave source unchanged. Source-pass endpoints differ from the original completion predicate.

Mandatory-proposal interfaces separate navigation from edit opportunity. Twelve exposed-task endings produce no applicable exact-string edit. A frozen line-span workflow on process close and an unused memory-stream issue~\cite{anyio728} accepts all 48 proposals and changes 20/24 final sources, but each arm scores 0/8. A second-model repeat accepts 22/24 proposals with each arm at 0/4. Active, method-specific source interventions and probes, expanded edit closure and mechanical multiline decoding accept 14/16 diagnostic edits and 23/24 repair proposals, yet again yield 0/4 per arm. An HTTPcore transfer also yields 0/2 per arm, with heavy diagnostic truncation. Interface-limited histories and complete costs remain in the supplement; these dependent development trials establish no explanation advantage and are not pooled.

The benefit supplement retains local gains, null results and contrary Skill transfer. Different endpoints and task pools preclude pooling; no deployed-folding or general Skill-evolution gain is established. \emph{Drafting assistance: OpenAI Codex~\cite{aiassist}.}

\section{Relation to prior work}
VeriSkill attributes verification failures, abstracts lessons and admits Skill revisions through executable validation while preserving the original executable program~\cite{veriskill}. We study changes to production behavior that also participates in validation. Our prediction and online-Skill experiments use different endpoints and do not implement VeriSkill's admission method. SkillEvolBench separates experience reuse from Skill formation through frozen transfer~\cite{skillevolbench}; SkillEvoReg uses perturbations and paired candidate-conditioned counterexamples~\cite{skillevoreg}. Counterexample admission is therefore prior work.

Trace-augmented repair and probing are established: Haque et al. report configuration-dependent gains~\cite{haquetraces}; TraceRepair combines probes and repair perspectives~\cite{tracerepair}. Our shared-observation controls test added interpretation, without reproducing or outperforming these systems.

ECK binds code to contracts, evidence, provenance and freshness~\cite{eck}; ContextProof versions source and dependencies~\cite{contextproof}. Both can represent stronger checks. Our contrasts concern which operation, context and interval an observation supports. Source-bound guidance is an inspired condition, not an ECK reproduction.

SkillsVote and AgentFactory govern Skill evolution and executable subagents~\cite{skillsvote,agentfactory}. BRIDGE studies provenance/counterfactual handoff repair and HSC-Bench state-dependent continuation~\cite{bridge,hscbench}; CTA, SkillTriage and Agent Skill Evolution examine trajectories, attribution and revisions~\cite{cta,skilltriage,skillevolution}. Our delivered revisions show no observed behavioral-transfer advantage, and fixed-checkpoint results do not identify deployed folding effects.

\emph{Whose Oracle Is It?} protects acceptance-witness dependencies while excluding production behavior under repair~\cite{oracleauthority}. Our command-dispatch cases place production code in both startup and target roles; substituting startup gives conditional evidence. BSG-VA binds commands to source states and replays bug contrasts~\cite{bsg}; our equal-triple counterexample does not refute its richer method. SWE-Doctor diagnoses pre-target failure and partial repair~\cite{swedoctor}. Building to the Test ablates unintended behavior suppliers~\cite{buildingtest}; our lifetime controls distinguish cleanup from live forwarding. Veritas and The Verification Horizon examine branch dependence and verifier faithfulness~\cite{veritas,verificationhorizon}.

Existing audits examine mocks, replay and phase integrity~\cite{mockbench,codetracer,benchshield,agentlenslucky,overmocked}, test-writing frequency~\cite{agenttests}, verifier restrictions~\cite{verifierbench}, harnesses~\cite{swtbench,specguard}, patch--issue alignment~\cite{retrace}, test validity/detection~\cite{testsuitebench} and amplification~\cite{dspot}. Plausibility, correctness, propagation and contamination also have established analyses~\cite{qiplausibility,smithoverfit,ahmedoverfit,ccdefects,ccmasri,manybugs,vibepass}. Our bounded dependency analysis and counterexamples establish neither a new provenance format nor superiority to these systems. \emph{Drafting assistance: OpenAI Codex~\cite{aiassist}.}

\section{Implications and limitations}
Evidence concerns an operation under stated source, context and lifetime conditions. Startup substitution changes context; cleanup and live flushing concern different intervals. Qualify both the actual default editor and protected checker.

Selection limits external validity: 118 original episodes cover 31 mechanism-relevant tasks and two model families; later Java/Rust studies reuse 12 development tasks. Handoff and C++ cases are neither representative nor necessarily pretraining-clean. Most later generated-note comparisons use Qwen; an exposed-task Devstral repeat also finds no full repair. Repeated seeds, inputs and library interventions remain within-task observations.

Many diagnoses/consumers are source-informed and post-hoc. Polling schedules are constructed, annotations use an application contract, and optional metadata lacks deployment support. Ordinary checks already detect all 12 H/I failures. Finite contrasts establish neither prevalence, diagnosis-time savings nor general repair superiority. A 26-call, one-revision Skill prediction probe initially worsens predictions 9/9 to 5/9; final-JSON-only scoring reverses its overall comparison. The supplement retains representation sensitivity, all failed qualifications and contrary controls. No deployed-folding or general evolution benefit follows. \emph{Drafting assistance: OpenAI Codex~\cite{aiassist}.}

\section{Conclusion}
Correct finite judgments can coexist with incorrect facts or excessive completion claims. Our contrasts locate how production edits change reachability, witness production and consumption; the Rust replay additionally separates registered work from admission in progress. Local repair differences do not establish diagnostic superiority, deployed-folding or general evolution gains. Bind claims to executed operations and intervals, verify evidence consumption, and assess repairs separately from their explanations. \emph{Drafting assistance: OpenAI Codex~\cite{aiassist}.}

\section*{Reproducibility}
Research archives retain inputs, outcomes, failures, recoveries and score/cost auditors for 118 original endpoints/860 calls and later cohorts. Their README maps denominators and verification commands. The experimental archive is currently maintained privately; public release is planned. Weights, runtime binaries and dependencies must be obtained separately to regenerate inference. \emph{Drafting assistance: OpenAI Codex~\cite{aiassist}.}

\section*{Acknowledgements}
OpenAI Codex assisted with drafting and revising text in the abstract, Sections~I--X, the Reproducibility section, and the supplementary material, and with preparing experimental and analysis code, tables, and figures~\cite{aiassist}.

\end{document}